\documentclass[runningheads]{llncs}
\usepackage{graphicx}
\usepackage{color}

\usepackage{enumitem}
\usepackage{amsmath}
\usepackage{xspace}

\usepackage{algorithm}
\usepackage[noend]{algpseudocode}

\begin{document}

\title{The Dynamic Travelling Thief Problem: Benchmarks and Performance of Evolutionary Algorithms}%\vspace{-3mm}}
\titlerunning{The Dynamic Travelling Thief Problem}
\author{Ragav Sachdeva, 
Frank Neumann, 
Markus Wagner}
%\author{Mahfouth Alghamdi\orcidID{0000-0002-0261-8002} \and
% Christoph Treude\orcidID{0000-xxxx} \and
% Markus Wagner\orcidID{0000-0002-3124-0061}}
%
\authorrunning{R. Sachdeva et al.}
\institute{School of Computer Science, University of Adelaide, Adelaide, Australia
\email{ragav.sachdeva@student.adelaide.edu.au}\\  \email{frank.neumann@adelaide.edu.au}, \email{markus.wagner@adelaide.edu.au}}%\vspace{-4mm}

\maketitle              % typeset the header of the contribution
\begin{abstract}
%\vspace{-1mm}
Many real-world optimisation problems involve dynamic and stochastic components.
While problems with multiple interacting components are omnipresent in inherently dynamic domains like supply-chain optimisation and logistics, most research on dynamic problems focuses on single-component problems.  
With this article, we define a number of scenarios based on the Travelling Thief Problem to enable research on the effect of dynamic changes to sub-components.
Our investigations of 72 scenarios and seven algorithms show that -- depending on the instance, the magnitude of the change, and the algorithms in the portfolio -- it is preferable to either restart the optimisation from scratch or to continue with the previously valid solutions. 

%Contributions: (1) We define the dynamic travelling thief problem (DynTTP) where the availability of items and cities can change over time, and (2) we propose and evaluate a number of problem-specific strategies that range from restarting from scratch to reoptimising from the previously best solution. 

\vspace{-2mm}
% \and combinatorial optimisation}
\vspace{-2mm}
\end{abstract}

\sloppy

%%%%%%%%%%%%%%%%%%%%%%%%%%%%%%%%%%%%%%%%%%%%%%%%%%%%%%%%%%%
%%%%%%%%%%%%%%%%%%%%%%%%%%%%%%%%%%%%%%%%%%%%%%%%%%%%%%%%%%%
%%%%%%%%%%%%%%%%%%%%%%%%%%%%%%%%%%%%%%%%%%%%%%%%%%%%%%%%%%%

\section{Introduction}

Real-world optimisation problems often involve dynamic and stochastic components, and evolutionary algorithms have shown to be very successful for dealing with such problems.This is due to the very nature of evolutionary computing techniques that allows them to adapt to changing environments without having to restart the algorithm when the problem characteristics change.
In terms of dynamic problems, the main focus has been one studying the behaviour of evolutionary algorithms for dealing with dynamic fitness functions. Here the dynamic setting is usually specified by altering the objective function dynamically and the goal is to re-track a moving optimal peak.

%\mw{maybe add a sentence or two on what's \emph{dynamic} -- the following jumps directly into constraints}

Recently, problems with dynamically changing constraints have been studied in the evolutionary computation literature both from a theoretical~\cite{DBLP:journals/corr/abs-1806-08547,DBLP:journals/algorithmica/ShiSFKN19} and experimental perspective~\cite{DBLP:conf/cec/Ameca-AlducinHB18,DBLP:conf/evoW/Ameca-AlducinHN18,DBLP:conf/ppsn/Roostapour0N18,DBLP:journals/corr/abs-1811-07806}. Here the objective function %to be optimised 
stays fixed and constraints change over time. This reflects scenarios where the goal is to maximise profit but the resources to achieve this goal can vary over time. A classical example are supply chain management problems where trucks and trains can become (un)available due to maintenance and break downs.  
Studies include important problems such as the knapsack problem~\cite{DBLP:conf/ppsn/Roostapour0N18} and more generally the optimisation of submodular functions~\cite{DBLP:journals/corr/abs-1811-07806} under dynamically changing constraints. In addition, for the knapsack problem, the optimisation under dynamic and stochastic constraints has also been investigated~\cite{DBLP:journals/corr/abs-2002-06766}.

In this paper, we investigate dynamic variants of the travelling thief problem (TTP). The TTP has been introduced in the evolutionary computation literature in \cite{DBLP:conf/cec/BonyadiMB13} as a multi-component problem combining the classical travelling salesperson problem (TSP) and the knapsack problem (KP). 
The TTP comprises a thief stealing items with weights and profits from a number of cities. The thief has to visit all cities once and collect items such that the overall profit is maximised. The thief uses a knapsack of limited capacity and pays rent for it proportional to the overall travel duration. To make the two components (TSP and KP) interdependent, the speed of the thief is made non-linearly dependent on the weight of the items picked so far.
The interactions of the TSP and the KP in the TTP result in a complex problem that is hard to solve by tackling the components separately. The TTP has gained significant attention in the evolutionary computation literature and several competitions organised to solve this problem have led to significant progress improving the performance of solvers.

In order to further extend the impact and significance of studies around the TTP, and to support research on the effects of dynamic changes to sub-components of problems, we design two types of dynamic benchmarks. The first one relates to the KP where items in cities can become available and unavailable. This inevitably changes the options of items that can be collected. %The third one is a combination of the first two \rs{third needs to be revisited here}.
The second one relates to the TSP component of the problem where cities can be made available and unavailable modelling the change in availability of locations.

We design different scenarios for these dynamic settings and examine how evolutionary algorithms based on popular algorithmic components for the TTP can deal with such changes. In particular, we are interested in learning when one should continue with the optimisation by recovering from the previous solution, or when one should create a new solution from scratch as a starting point for the next iterative improvement.

%The remainder of this article is organised as follows. \mw{...}
%In dynamic problems, the question is always: should one attempt to salvage from the previously valid solution what is possible (i.e., to consider it as a starting point), or should one approach the changed situation from scratch?

%%%%%%%%%%%%%%%%%%%%%%%%%%%%%%%%%%%%%%%%%%%%%%%%%%%%%%%%%%%
%%%%%%%%%%%%%%%%%%%%%%%%%%%%%%%%%%%%%%%%%%%%%%%%%%%%%%%%%%%
%%%%%%%%%%%%%%%%%%%%%%%%%%%%%%%%%%%%%%%%%%%%%%%%%%%%%%%%%%%

\section{The Travelling Thief Problem and its Variants}

%%%% DROPPED on 200628 FOR ICONIP
The TTP is a combinatorial optimisation
problem that aims to provide testbeds for solving problems with multiple interdependent components~\cite{DBLP:conf/cec/BonyadiMB13,BenchmarkPaper}. 
In the following, we first introduce the TTP. Then, we put this present research into the context of existing works by reviewing existing TTP variants. 
\subsection{Definition}\label{ref:ttpdef}
%%%%%%%%%%%%%%%%%%%%%%

The TTP combines two well-known problems, namely, the Travelling Salesman Problem and the Knapsack Problem.

The problem is defined as follows: we are given a set of
$n$ cities, the associated matrix of distances $d_{ij}$, and
a set of $m$ items distributed among these cities. Each item $k$ is defined by
a profit $p_{k}$ and a weight $w_k$. A thief must visit all the cities exactly once, stealing
some items on the road, and return to the starting city.

The knapsack has a capacity limit of $W$, i.e. the total weight of the collected items must not exceed $W$.
In addition, we consider a renting rate $R$ that the thief must pay at
the end of the travel, and the maximum and minimum velocities denoted
$v_{max}$ and $v_{min}$ respectively. Furthermore, each item is available in only one city,
and $A_i \in \{1, \dots, n\}$ denotes the availability vector. $A_i$ contains
the reference to the city that contains the item $i$.

A TTP solution is typically coded in two parts: the tour $X = (x_1, \dots, x_n)$,
a vector containing the ordered list of cities, and the picking plan
$Z = (z_1, \dots, z_m)$, a binary vector representing the states of items
($1$ for packed, $0$ for unpacked).

To establish a dependency between the sub-problems, the TTP was designed such that the speed of
the thief changes according to the knapsack weight. To achieve this, the thief's velocity at
city $c$ is defined as 
% in Equation~\ref{eq:vel}.
% \begin{equation}
% \label{eq:vel}
$v_x = v_{max} - C \times w_x$, 
% \end{equation}
where $C = \frac{v_{max}-v_{min}}{W}$ is a constant value, and $w_x$ is the weight of the knapsack
at city~$x$.

% We note $g(z)$ the total items value (defined in Equation~\ref{eq:items}), 
% and we note $f(x, z)$ the total travel time (defined in Equation~\ref{eq:time}).
%\begin{align} \label{eq:items}
The total value of items is
$g(Z) = \sum_m p_m \times z_m$, such that $\sum_m w_m \times z_m \le W $.
%\end{align}
The total travel time is
%\begin{equation} \label{eq:time}
$f(X, Z) = \sum_{i=1}^{n-1} t_{x_i, x_{i+1}} + t_{x_n, x_1}$, 
%\end{equation}
where $t_{x_i, x_{i+1}} = \frac{d_{x_i, x_{i+1}}}{v_{x_i}}$ is the travel time from $x_i$ to $x_{i+1}$.

The TTP's objective is to maximise the total travel gain function, 
which is the total profit of all items minus the travel time multiplied with the renting rate: 
% as defined in Equation~\ref{eq:gain},
% by finding the best tour and picking plan.
% \begin{equation}
% \label{eq:gain}
$F(X, Z) = g(Z) -  f(X, Z) \times R$.
% \end{equation}

For a worked example, we refer the interested reader to the initial TTP article by Bonyadi et al.~\cite{bonyadi2013travelling}.

%%%% DROPPED on 200628 FOR ICONIP
\subsection{TTP variants to date}

Since the inception of the TTP by Bonyadi et al.~\cite{bonyadi2013travelling}, research has focused not only on the original formulations, but also on a few variants have been created to either address its shortcomings or to investigate the interaction of the components. In the following we provide a brief overview.

Thus far, most research has considered the single-objective TTP formulation (TTP1) from \cite{bonyadi2013travelling}, which is typically the TTP variant that is referred to as \emph{the TTP} (see Section~\ref{ref:ttpdef}). The other inaugural variant (TTP2) considers two objectives and additionally, a value drop of items over time.

For the single-objective TTP, a wide range of approaches has been considered, ranging from general-purpose iterative search-heuristics~\cite{BenchmarkPaper}, co-evolutionary approaches~\cite{bonyadi2014socially,namazi2019cooperative} and memetic approaches~\cite{mei2014interdependence} to swarm-intelligence based approaches~\cite{wagner2016stealing,Zouari2019antstpp}, to approaches with problem specific search operators~\cite{faulkner2015approximate}, 
%On a higher, i.e., algorithmic level, 
customised estimation of distribution approaches \cite{Martins2017ttpeda} and to hyper-heuristics~\cite{elyafrani2018hyperttp}. % have been explored.
Wagner et al.~\cite{wagner2018casestudy} provide a comparison of 21 algorithms for the purpose of portfolio creation. 
To better understand the effect of operators %on the capability to find good solutions 
on a more fundamental level, \cite{elyafrani2018ttplandscape} and \cite{Wuijts2019ttpinvest} present, through fitness-landscape analyses, correlations and characteristics that are potentially exploitable.

For a variant with fixed tours, named Packing-While-Travelling (PWT), 
Neumann et al.~\cite{neumann2017ttpPTAS} %created a subproblem of the TTP called Packing-While-Travelling (PWT), which is equivalent to the TTP but the tours remain fixed. For this variant, they 
showed that it can be solved in pseudo-polynomial time via dynamic programming, by exploiting the fact that the weights are integer.  \cite{wu2017ttpexact} extended this to optimal approaches for the entire TTP, however, their approaches are only practical for very small instances.

As the TTP's components are interlinked, multi-objective considerations that investigate the interactions via the idea of ``trade-off''-relationships are becoming increasingly popular as of lately. 
For example, \cite{yafrani2017ttpemo} created an approach 
%that generates diverse sets of solutions, while being competitive with the state-of-the-art single-objective algorithms; 
where the objectives were the travel time and total profit of items. 
\cite{wu2018evolutionary} considered a bi-objective version of the TTP, where the objectives were the total weight and the TTP objective score. 
%This hybrid approach makes use of the dynamic programming approach for the PWT and then searches over the space of tours only. 
At two recent competitions\footnote{\textit{EMO-2019}, \url{https://www.egr.msu.edu/coinlab/blankjul/emo19-thief/} and \textit{GECCO-2019} \url{https://www.egr.msu.edu/coinlab/blankjul/gecco19-thief/}}, an intermediate version of TTP1 and TTP2 was proposed, %. The problem description 
which can be seen as a TTP1 with two objectives or TTP2 without a value drop of items over time. The same TTP variant was investigated by \cite{blank2017solvingBittp}.
to build the bridge from TTP1 to TTP2 by having two objectives but not adding another level of complexity to the problem. 
%The proposed problem definition aimed to build the bridge from TTP1 to TTP2 by having two objectives but not adding another level of complexity to the problem. 
Recently, Chagas~et~al.~\cite{chagas2020nondominated} proposed a NSGA-II and customised random-key genetic algorithm for this problem.

Lastly, %although the TTP is a challenging problem, 
it is often argued if the TTP is realistic enough because of its formulation -- not because it is static, which we are addressing in this article -- but because it requires a single thief to travel across hundreds or thousands of cities to steal items. %In addition, the thief is forced to visit all cities, regardless of whether an item is stolen there or not. 
Chand et al.~\cite{chand2016fast}'s Multiple Travelling Thieves Problem relaxes several constraints so that a group of thieves can maximise the group's collective profit by going on different tours and without the need to visit all cities.

A more general discussion of a multi-objective approach to interconnected problems can be found in~\cite{klamroth2016interwoven}, and a more general discussion on the opportunities of multi-component problems can be found in~\cite{Bonyadi2019}.

Note that our research in this article has been conducted in parallel to and independently of that presented by Herring~et~al.~\cite{herring2020dynamic} on multi-objective dynamic TTP variants. There, the authors considered dynamic city locations, dynamic item availability and dynamic item values. Experimentally, they considered one dynamic setting for several small TTP instances, and they did not investigate the effects of disruptions in isolation.

% \subsection{Dynamic Problems}

% \mw{Is the DynTTP the first dynamic multi-component problem? What is the most similar? Again, mention that the TTP is an academic problem that let's us study effects in relatively pure settings.}

%%%%%%%%%%%%%%%%%%%%%%%%%%%%%%%%%%%%%%%%%%%%%%%%%%%%%%%%%%%
%%%%%%%%%%%%%%%%%%%%%%%%%%%%%%%%%%%%%%%%%%%%%%%%%%%%%%%%%%%
%%%%%%%%%%%%%%%%%%%%%%%%%%%%%%%%%%%%%%%%%%%%%%%%%%%%%%%%%%%

% \section{Framework}

% In the following, we first define the Dynamic Travelling Thief Problem

\section{Dynamic Travelling Thief Problem (DynTTP)}

Broadly speaking, DynTTP is an extension of the classic TTP where the problem constraints can change during run-time. In this study, we investigate the introduction of dynamism in two different ways by flipping the availability status of $d\%$ of the (1) items and (2) cities uniformly at random, i.e. if an item/city is available, it is made unavailable and vice-versa. From here on, we will call the event when availabilities change a \emph{disruption}, and the time span between two disruptions is called an \emph{epoch} which is of duration $z$.

To allow for a meaningful investigation, both these dynamic changes are studied independently of one another. Furthermore, while ``Toggling Items'' the tour is kept fixed and similarly while ``Toggling Cities'' the packing list is not re-optimised. In addition, to guarantee that the solutions remain valid and to permit a fair analysis, we make the following design decisions:

\paragraph{Toggling Items:}
If an item is made unavailable, it is no longer in the current packing plan (if already picked) and cannot be picked again by the thief.
On the other hand, if an item is made available again, it is available for the thief to steal but it is not automatically added to the current packing list (to prevent unintentionally exceeding the knapsack capacity).

\paragraph{Toggling Cities:}
If a city is made unavailable, it is removed from the current tour while maintaining the order of the remaining cities. Of course, this also implies that the items in this city are no longer available either.
On the other hand, if a city is made available again, it is inserted back into the tour at the same position as it was before.
Furthermore, toggling a city back in, also updates the knapsack to include the items from this city that were previously in the packing list. This is an acceptable thing to do because even if all the cities are toggled `on', the knapsack will not exceed capacity as the packing plan would revert back to the initial solution at $t=0$.

\begin{figure}[t]
\centering\vspace{-2mm}
  \includegraphics[width=95mm]{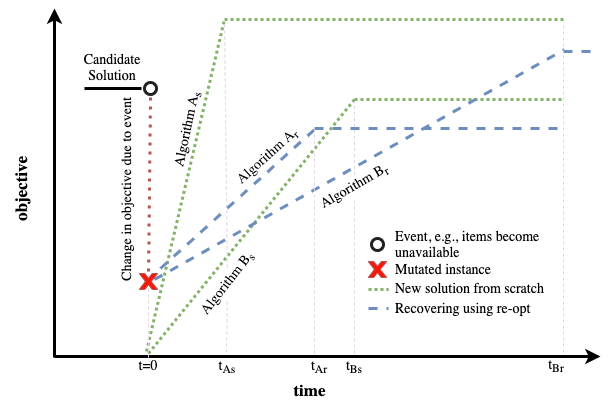}\vspace{-3mm}
  \caption{Simplified DynTTP scenario: as time $t=0$, an instance gets disrupted, and a decision maker has to decide (1) whether to employ algorithm $A$ or algorithm $B$ and (2) whether to recover from the previous solution $R$ 
%(marked \raisebox{-0.5mm}{{\includegraphics[width=3mm]{redx}}})
(marked \raisebox{-0.1mm}{\color{red}\textbf{\textsf{X}}}) 
or to optimise from scratch $S$. $t_{AS} \ldots t_{BR}$ denote the times needed by the different choices to converge.}\vspace{-2mm}
  \label{fig:motiv}
\end{figure}

\vspace{1.5mm}We show a simplified DynTTP scenario in Figure~\ref{fig:motiv}. As it is often the case with dynamic optimisation, it depends on the magnitude of the change, on the available computational budget, and on the available algorithms whether it is preferable to restart the optimisation from scratch or to recover based on the previous solution.

% \paragraph{Implementation note: }
\label{ref:randomnumbers} 
Note that a dedicated random number generator is responsible for modifying the instances. This way, different algorithms observe the same set of conditions for a given instances, and thus we ensure that sequences of disruptions remain comparable, as otherwise instances would drift apart and the algorithms' performances become incomparable.

%%%%%%%%%%%%%%%%%%%%%%%%%%%%%%%%%%%%%%%%%%%%%%%%%%%%%%%%%%%
%%%%%%%%%%%%%%%%%%%%%%%%%%%%%%%%%%%%%%%%%%%%%%%%%%%%%%%%%%%
%%%%%%%%%%%%%%%%%%%%%%%%%%%%%%%%%%%%%%%%%%%%%%%%%%%%%%%%%%%

\section{Heuristics for the DynTTP}\label{sec:algorithms}

As the items/cities are toggled ensuing from the dynamic change, the objective value of the known TTP solution changes. At this point there are two choices: Re-build a TTP solution from scratch for, what is effectively a new TTP instance; or re-optimise the packing/tour for the previously known best solution.

To achieve this, we make use of a number of generic as well as problem-specific algorithmic building blocks.

\newcommand\bitflip[0]{\textsc{Bitflip}\xspace}
\newcommand\packit[0]{\textsc{PackIterative}\xspace}
\newcommand\packitbitflip[0]{\textsc{PackIterative+Bitflip}\xspace}
\newcommand\insertion[0]{\textsc{Insertion}\xspace}
\newcommand\clk[0]{\textsc{CLK}\xspace}
\newcommand\clkinsertion[0]{\textsc{CLK+Insertion}\xspace}
\newcommand\rea[0]{\textsc{REA$_m$}\xspace}

\bitflip~\cite{ApproximateApproaches} is a simple greedy hill-climber with a low runtime. It iteratively evaluates the outcome of flipping each bit position corresponding to item $i$ in the packing plan $Z$. If a flip improves the objective value, the change is kept, otherwise $Z$ remains unchanged. A single iteration of the operator ends when all bit flips have been attempted once. Multiple passes can result in further improvement of the objective value.

\packit~\cite{ApproximateApproaches} is a greedy packing heuristic with a scoring function that computes a trade-off between a distance that item $i$ has to be carried over, its weight and its profit. As it might be beneficial to change the influence of the distance %strengthen the influence of one or more variables of the function 
depending on the particular instance, \packit performs an interval search over one parameter that balances the respective influences. %Thus, \packit consumes more than one evaluation. 
    
\insertion~\cite{ApproximateApproaches} is another simple greedy hill-climber, and very similar to \bitflip. It operates on tours and takes advantage of the situation where a valuable item at a particular city is picked up early and it is worth trying to delay the item pickup by modifying the tour. 
    
\clk (Chained Lin-Kernighan heuristic~\cite{chainedLK03applegate}\footnote{As available at \url{http://www.tsp.gatech.edu/concorde/downloads/downloads.htm}}) is a fast TSP solver. While research optimal TTP tours are not necessarily TSP-optimal, it is nevertheless a popular starting point in TTP research.
    
\rea is a variant of the simple population-based ($\gamma$+1)EA outlined by Doerr et al.~\cite{ReoptViaStructuralDiversity} that utilises a diversity mechanism to re-optimise the solution to a perturbed instance. It works with a diverse set of solutions at Hamming distance of at most $\gamma$ from a previously known solution $x^{old}$. %, where $\gamma$ is a parameter of the algorithm. 
%In addition, the algorithm maintains the best found solution $x^{best}$ thus far.
In every iteration, the algorithm first selects a parent individual $x$, from the diverse set of solutions along with the current best solution, via a biased random selection. Following this, one offspring $y$ is generated using a standard bit mutation with the rate of $1/n$. If the Hamming distance $i = H(y, x^{old})$ is at most $\gamma$, this offspring $y$ replaces the previous best individual $x^{i}$ at distance $i$ if it is at least as good.
It is imperative to point out that the ``change event'' studied by Doerr et al. is conceptually different to what we are investigating: in their work %is concerned with the situation where 
the objective function $f^{old}$ is perturbed resulting in a new objective function $f$ and their problem definition imposes no condition where their known solution becomes invalid after the change. To adapt to these fundamental differences, we employ this algorithm where the initial known solution i.e. $x^{old}$ is the resulting solution \textit{after} the disruption and $\gamma$ is set to $m$ (number of items originally available). The merit of using this diversity mechanism is to strike a balance between exploring the neighbourhoods of a previously known good solution vs. hill-climbing the best solution thus far.

% \begin{algorithm}
% \caption{REA$_m$}\label{alg:REAModified}
% \begin{algorithmic}[1]
% \State \textbf{Input:} Solution $x^{mutated}$ 
% \State \textbf{Initialisation:}
% \State \hskip1em $x^{best} \gets x^{mutated}$
% \State \hskip1em \textbf{for} $i=0,1,2...,n-1$ \textbf{do}  $x^{i} \gets x^{mutated}$
% \State \textbf{Optimisation: while} within time budget \textbf{do}
% \State \hskip1em with a probability of $1/2$:
% \State \hskip1.5em $x^{c} \gets x^{best}$ or;
% \State \hskip1.5em $x^{c} \gets$ randomly from $\{x^{i}| i\in [0..n-1]\}$
% \State \hskip1em Create $y$ from $x^{c}$ by flipping in each bit 
% \State \hskip1.5em independently with probability $1/n$;
% \State \hskip1em \textbf{if} $Z(y)\geq Z(x^{best})$ \textbf{then} $x^{best}\gets y$
% \State \hskip1em $h \gets$ HammingDistance($y$, $x^{mutated}$)
% \State \hskip1em \textbf{if} $Z(y)\geq Z(x^{h})$ \textbf{then} $x^{h}\gets y$
% \end{algorithmic}
% \end{algorithm}

%\mw{to decide if we need abbreviations here, i.e., I1..I4 and C1..C3}

For the different scenarios of the dynamically changing availability of items and cities, we combine the building blocks in the following seven ways.\label{ref:algorithm}

\vspace{2mm}
\noindent\textit{Toggling Items -- four approaches:}
\vspace{-2.mm}
\begin{enumerate}
    \item Re-optimise the mutated packing using \bitflip;
    \item Re-optimise the mutated packing using \rea with the population size of $m$, which is the number of items originally available.
    \item Create a new packing plan from scratch using \packit;
    \item Create a new packing plan from scratch using \packit and then optimise it using \bitflip;
\end{enumerate}

%\pagebreak%\vspace{2mm}
\noindent\textit{Toggling Cities -- three approaches:}
\vspace{-2.mm}
\begin{enumerate}
    \item Re-optimise the mutated tour using \insertion;
    \item Create a new tour from scratch using \clk;
    \item Create a new tour from scratch using \clk and then optimise it using \insertion.
\end{enumerate}

\section{Computational Investigation}
\label{sec:experiments}

In the following, we first describe our experimental setup and the 72 scenarios. Then, we present the results and highlight situations that we find interesting.

We have made all the code, the results, the and processing scripts publicly available at \emph{\url{https://github.com/ragavsachdeva/DynTTP}}. This includes cross-validated C++ implementations of the algorithms. %\bitflip, \insertion, and \packit. 

\subsection{Experimental setup}

Polyakovskiy et al. presented a systematically created set of benchmark TTP instances that cover a wide range of features and are based on well known instances of TSP and KP \cite{BenchmarkPaper}. The following set of TTP instances covers a range of scenarios for the travelling thief, and therefore was selected for the purposes of this study:\footnote{It has also been the subset used at the 2014/2015 and 2017 TTP competitions, e.g., \url{https://cs.adelaide.edu.au/~optlog/TTP2017Comp/}}

\newcommand\asmall[0]{A279\xspace}
\newcommand\amedium[0]{A1395\xspace}
\newcommand\alarge[0]{A2790\xspace}
\newcommand\fsmall[0]{F4460\xspace}
\newcommand\fmedium[0]{F22300\xspace}
\newcommand\flarge[0]{F44600\xspace}
\newcommand\psmall[0]{P33809\xspace}
\newcommand\pmedium[0]{P169045\xspace}
\newcommand\plarge[0]{P338090\xspace}

\vspace{2mm}\noindent {\footnotesize
\begin{tabular}{ll}
280\textunderscore n279\textunderscore bounded-strongly-corr\textunderscore 01 
& a280\textunderscore n1395\textunderscore uncorr-similar-weights\textunderscore 05 \\
a280\textunderscore n2790\textunderscore uncorr\textunderscore 10
& fnl4461\textunderscore n4460\textunderscore bounded-strongly-corr\textunderscore 01 \\
fnl4461\textunderscore n22300\textunderscore uncorr-similar-weights\textunderscore 05
& fnl4461\textunderscore n44600\textunderscore uncorr\textunderscore 10 \\
pla33810\textunderscore n33809\textunderscore bounded-strongly-corr\textunderscore 01
& pla33810\textunderscore n169045\textunderscore uncorr-similar-weights\textunderscore 05 \\
pla33810\textunderscore n338090\textunderscore uncorr\textunderscore 10
\end{tabular}
}\vspace{2mm}

% \vspace{-1mm}
% \begin{itemize}
%     \item a280\textunderscore n279\textunderscore bounded-strongly-corr\textunderscore 01 
%     \item a280\textunderscore n1395\textunderscore uncorr-similar-weights\textunderscore 05 
%     \item a280\textunderscore n2790\textunderscore uncorr\textunderscore 10
%     \item fnl4461\textunderscore n4460\textunderscore bounded-strongly-corr\textunderscore 01
%     \item fnl4461\textunderscore n22300\textunderscore uncorr-similar-weights\textunderscore 05
%     \item fnl4461\textunderscore n44600\textunderscore uncorr\textunderscore 10
%     \item pla33810\textunderscore n33809\textunderscore bounded-strongly-corr\textunderscore 01
%     \item pla33810\textunderscore n169045\textunderscore uncorr-similar-weights\textunderscore 05
%     \item pla33810\textunderscore n338090\textunderscore uncorr\textunderscore 10
% \end{itemize}

\vspace{-1mm}The numbers in each instance name are (from left to right) the total number of cities, the total number of items, % and the number of items per city (except the starting city, which has no items) respectively, 
and the knapsack size in that instance. For example, instance a280\textunderscore n2790\textunderscore uncorr\textunderscore 10 has 280 cities all of which (except the first) have 10 items each, totalling 2790 items, and it has the largest knapsack. In addition the terms uncorrelated, uncorrelated-similar-weights and bounded-strongly-correlated reflect the unique knapsack types as described in \cite{BenchmarkPaper}. 
To facilitate readability, we will use the abbreviations \asmall, \amedium, \ldots, \plarge, which include the number of items in their names.

In our computational study of the DynTTP, we investigate disruptions of the following types:
%Subject to our computational study of the DynTTP are the following components:

%\begin{itemize}
    %\item nine TTP instances, as defined above;
    %\item seven algorithms (see Section~\ref{sec:algorithms});
    %\item disruptions:
\vspace{-1mm}\begin{itemize}
        \item instance feature $f \in \left\{ \textrm{items, cities} \right\}$; %, items\&cities;
        \item amount of disruption $d \in \left\{ \textrm{1\%, 3\%, 10\%, 30\%} \right\}$;
        \item period $z$, which is the time between disruptive events. %, where $z$ is the number of items in the original instance. % \mw{if possible to $z^{1.5}$ and $z^2$ as well, and we see later if we can afford to spend a second page on one of them}.
    \end{itemize}\vspace{-1mm}
%\end{itemize}

As the disruptions to instances are random (modulo the determinism described in Section~\ref{ref:randomnumbers}), we perform 30 independent runs of each scenario -- a ``scenario'' is an instance, together with an instance feature $f$ that is to be disrupted, an amount of disruption $d$, and a period $z$.

We conduct all experiments on the university's compute cluster, which is equipped with Intel Xeon Gold 6148 processors (2.4 GHz). The initial solutions are created by running \clk first to generate a good TSP tour and then by running \packitbitflip to converge to a good TTP solution.

\subsection{A single disruptive event}

In our first study, we investigate how well the different algorithms perform given different scenarios. 
This study comprises 252 experiments, as we run 7 algorithms on the 9 instances that undergo 4 different amounts of disruption, and as mentioned above, we repeat each experiment 30 times. Following the disruption, each algorithm has a maximum of $z=10$ minutes, however, it can stop early if it detects convergence based on its internal criterion.

\begin{figure}[tbp]
\hspace*{50mm}(a) Toggling of items\vspace{2mm}\\%\line(1,0){347}\\
\hspace*{19mm} $d=3\%$ \hspace{26mm} $d=10\%$ \hspace{26mm} $d=30\%$\\
\vspace{-4mm}\rotatebox{90}{\hspace{15mm}\textsf{\scriptsize \bitflip }}
\includegraphics[width=39mm,trim=20 0 80 0,clip]{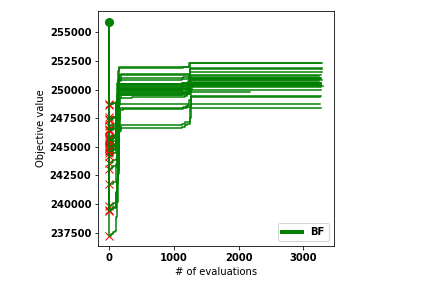} 
\includegraphics[width=39mm,trim=20 0 80 0,clip]{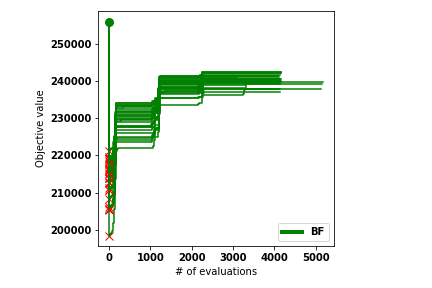}
\includegraphics[width=39mm,trim=20 0 80 0,clip]{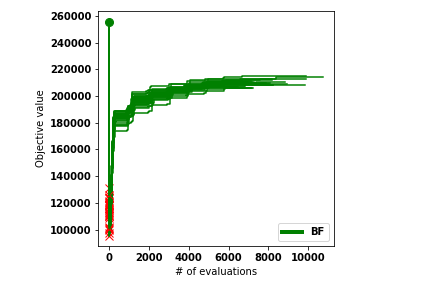}\vspace*{3mm}\\
\rotatebox{90}{\hspace{2mm}\textsf{\scriptsize \packit + \bitflip }}
\includegraphics[width=39mm,trim=20 0 80 0,clip]{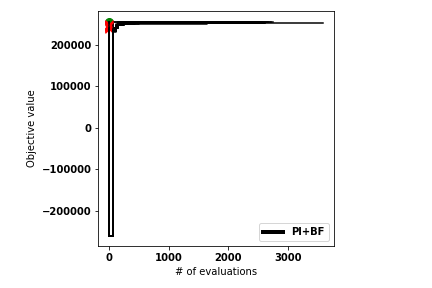}
\includegraphics[width=39mm,trim=20 0 80 0,clip]{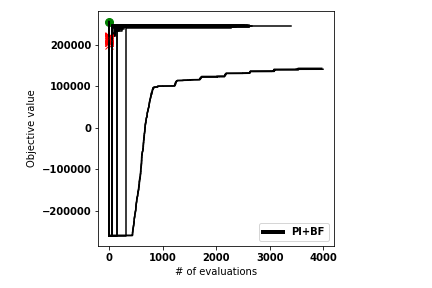}
\includegraphics[width=39mm,trim=20 0 80 0,clip]{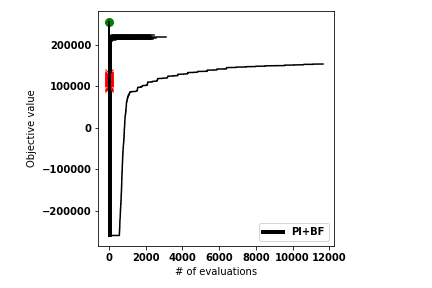}\\
% \vspace{-2mm}\caption{The availability of items changes. 
% Scenario: instance fnl4461\textunderscore n4460\textunderscore bounded-strongly-corr\textunderscore 01 \textunderscore items. 
% Optimised using two approaches: recovering from the previous solution (top row), starting from scratch (bottom row). 
% }
% \label{fig:singleDisruptionItems}\vspace{-3mm}
% \end{figure}

% \begin{figure}[tbp]
% \vspace{-1mm}
\hspace*{50mm}(b) Toggling of cities\vspace{2mm}\\%\line(1,0){347}\\
\hspace*{19mm} $d=3\%$ \hspace{26mm} $d=10\%$ \hspace{26mm} $d=30\%$\\
\vspace{-4mm}\rotatebox{90}{\hspace{13mm}\textsf{\scriptsize \insertion }}
\includegraphics[width=39mm,trim=20 0 80 0,clip]{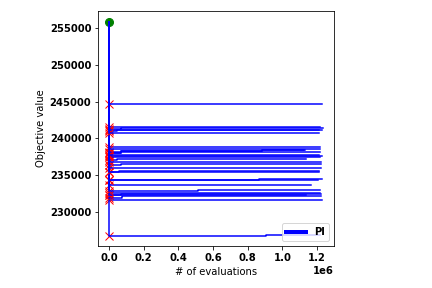}
\includegraphics[width=39mm,trim=20 0 80 0,clip]{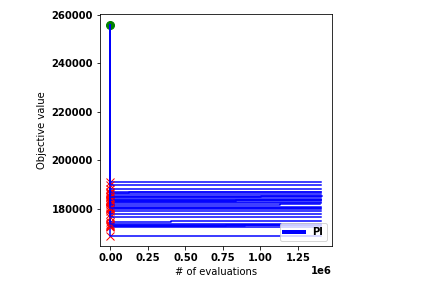}
\includegraphics[width=39mm,trim=20 0 80 0,clip]{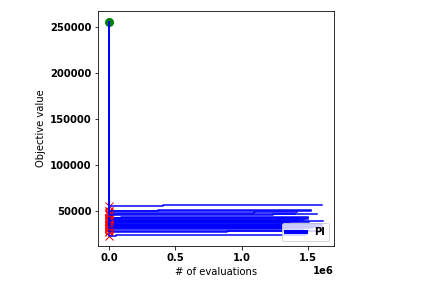}\vspace*{3mm}\\
\rotatebox{90}{\hspace{8mm}\textsf{\scriptsize \clkinsertion }}
\includegraphics[width=39mm,trim=20 0 80 0,clip]{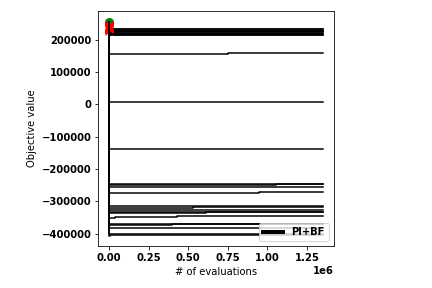}
\includegraphics[width=39mm,trim=20 0 80 0,clip]{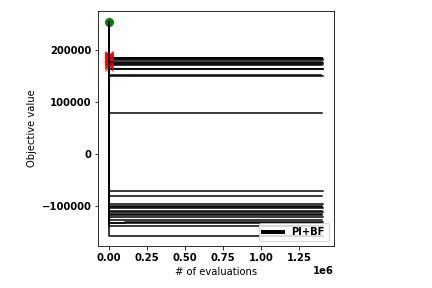}
\includegraphics[width=39mm,trim=20 0 80 0,clip]{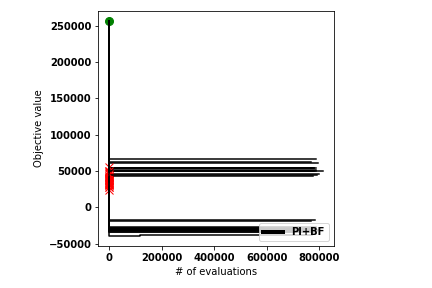}
\vspace{-2mm}\caption{DynTTP results of four algorithms on six scenarios for \fsmall. 
\textbf{Part (a)} Toggling of items: recovering from the previous solution (\bitflip, 1st row), starting from scratch (\packitbitflip, 2nd row). 
\textbf{Part (b)} Toggling of cities: recovering from the previous solution (\insertion, 1st row), starting from scratch (\clkinsertion, 2nd row).
Note that the scales have not be normalised on purpose to show the characteristics of the individual algorithms. 
Red crosses show the solution quality of the last solution given the new situation (see Figure~\ref{fig:motiv}), hence they represent the starting points for \bitflip and \insertion.
%For a normalised view at a larger study, please see Figure~\ref{fig:heatmapsAll}.
%The availability of cities changes. 
%Scenario: instance fnl4461\textunderscore n4460\textunderscore bounded-strongly-corr\textunderscore 01 \textunderscore items. 
%Optimised using two approaches: recovering from the previous solution (top row), starting from scratch (bottom row). 
}
\label{fig:singleDisruptionCities}\vspace{-3mm}
\end{figure}

In the following, we limit ourselves to subsets that show interesting situations. In general, we can typically make the following qualitative observations:
\vspace{-2mm}\begin{enumerate}
    \item \textit{Toggling Items:} When recovering from the previous solution, \bitflip makes only a small number of improvements. In Figure~\ref{fig:singleDisruptionCities} Part (a), we can see that there is only a small number of improvements for each of the 30 runs, even though 4460 bits can be flipped in the packing plan. When starting from scratch using \packitbitflip, we regularly observe a subset of runs that had difficulties reaching the quality of the solution prior to being discarded by \packit (see the red crosses in the staircase plots); hence, restarting from scratch can be seen as a risky choice here. 
    \item \textit{Toggling Cities:} When recovering from the previous solution, \insertion makes only a small number of improvements here as well; moreover, the improvements are very minor. We can see in Figure~\ref{fig:singleDisruptionCities} Part (b), that \clkinsertion has significantly more difficulties surpassing the previous solutions (red crosses) than the conceptually similar \packitbitflip has in the \textit{Toggling Items} scenarios.
    \item \rea, being a population-based approach for \textit{Toggling Items} scenario, often makes many improvements over the starting points. While it typically hits the time limit of 10 minutes (and amounting, e.g., 70,000 fitness evaluations on \fsmall and up to 10 million fitness evaluations on \asmall during this time), the trajectories imply that it was typically far from converging.
    \item \clk without a subsequent hillclimber typically performs badly for \textit{Toggling Cities}, as it is able to generate a TSP tour only once by a TSP solver, which ignores the fixed KP component. In contrast to this, \packit without a subsequent hillclimber performs significantly better as it performs an internal local search that attempts to maximise the TTP score.
\end{enumerate}

%to decide: include the items+cities toggles here as well?

\subsection{Multiple disruptive events}

Next, we investigate the performance over time for multiple disruptions. In particular, we observe 10 epochs and each lasts for a period of $z=m$ function evaluations (equal to the number of items in the original instance) or for 10 minutes, whichever comes first.
%where $m$ is the number of items available in the original instance. 
Again, we investigate 72 scenarios in total. Due to the high dimensionality of this study, (1) we use heatmaps to allow for a qualitative inspection of the results, and (2) we cut through the data in different ways and use statistical tests to compare the heuristic approaches.

\paragraph{Heatmaps.} 

\begin{figure}[t]
\centering\vspace{-3mm}
\rotatebox{90}{\hspace{7mm}\textsf{\scriptsize Objective value}}
\includegraphics[width=39mm,trim=30 0 0 0,clip]{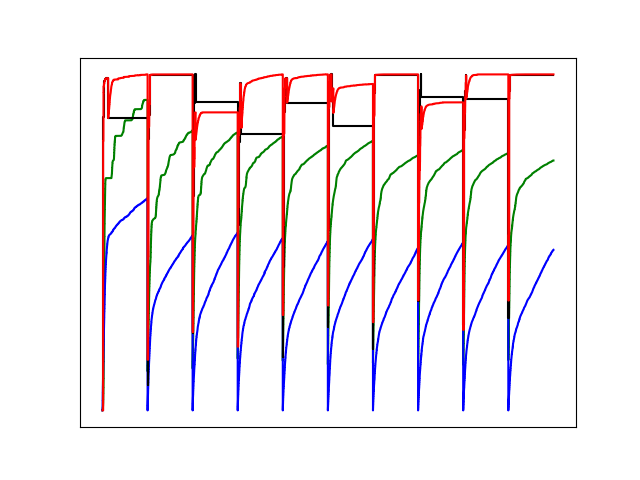}
\raisebox{13mm}{\Large$\Rightarrow$}
\hspace{3mm}
\rotatebox{90}{\hspace{9mm}\textsf{\scriptsize algorithms}}
\includegraphics[width=39mm,trim=30 0 0 0,clip]{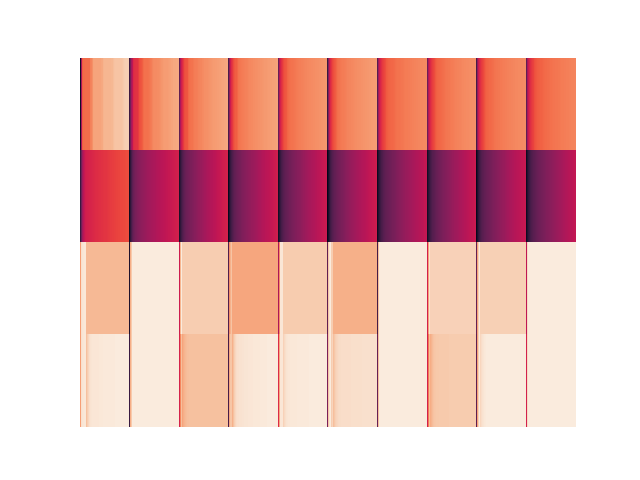}
\vspace{-4mm}\caption{\textit{Average} performance of four algorithms for 10 epochs (\fsmall, $d=30\%$). Time, as measured in the number of evaluations, passes from left to right; each period is of length $z=4460$ evaluations. Within each epoch, the performance is normalised across all algorithms. The lighter the colour, the higher the performance. Algorithms: \bitflip (green, 1st row), \rea (blue, 2nd row), \packit (black, 3rd row), \packitbitflip (red, 4th row). }
\label{fig:heatmapExample}\vspace{-2mm}
\end{figure}

Before moving on, we briefly explain the creation of each heatmap. First, we record for each of the 30 runs all TTP objective scores. Second, for each scenario and each epoch (as these are identical) we determine the minimum and maximum objective scores across all algorithms, and then linearly normalise the average of the 30 trajectories into $[0,1]$. These scores are then visualised as coloured bars, from black (lowest value seen by any algorithm in this scenario and epoch), via red to a light peach (highest value seen by any algorithm). 
Figure~\ref{fig:heatmapExample} shows an example for four algorithms. To facilitate readability, we omit all labels, because the period is fixed (as defined above), and because the order of the four (respectively three) approaches is stated in the captions.\footnote{The algorithm's order is also identical to the order listed in Section~\ref{ref:algorithm}.}

% \mw{todo cleanup and check for consistency}
% matrix of heatmaps: 
% \begin{itemize}
%     \item each heatmap contains all algorithm for the scenario
%     \item matrix: 9 columns (9 instances) X 12 rows (features X disruption)
%     \item heatmaps without x/y labels/ticks and without the z-axis colour bar$\rightarrow$ I'll arrange them (this will fill up one page)
% \end{itemize}

% \mw{The frequency was fixed to the number of items of the original instance (we do not use the number of cities).}

% \newcommand\hm[1]{\includegraphics[width=13mm,height=5mm,trim=58 38 46 42,clip]{heatmaps_multiMutations/#1}}
% \newcommand\hmFour[1]{%
% \hm{#1=1.png}
% \hm{#1=3.png}
% \hm{#1=10.png}
% \hm{#1=30.png}%
% }

\newcommand\hmOneOfNine[1]{\includegraphics[width=11.5mm,height=14mm,trim=58 38 46 42,clip]{heatmaps_multiMutations/#1} }%
\newcommand\hmNine[2]{%
\hmOneOfNine{a280_n279_bounded-strongly-corr_01_#1_k=#2.png}%
\hmOneOfNine{a280_n1395_uncorr-similar-weights_05_#1_k=#2.png}%
\hmOneOfNine{a280_n2790_uncorr_10_#1_k=#2.png}%
\hmOneOfNine{fnl4461_n4460_bounded-strongly-corr_01_#1_k=#2.png}%
\hmOneOfNine{fnl4461_n22300_uncorr-similar-weights_05_#1_k=#2.png}%
\hmOneOfNine{fnl4461_n44600_uncorr_10_#1_k=#2.png}%
\hmOneOfNine{pla33810_n33809_bounded-strongly-corr_01_#1_k=#2.png}%
\hmOneOfNine{pla33810_n169045_uncorr-similar-weights_05_#1_k=#2.png}%
\hmOneOfNine{pla33810_n338090_uncorr_10_#1_k=#2.png}%
\vspace{-2mm}
}

\begin{figure}
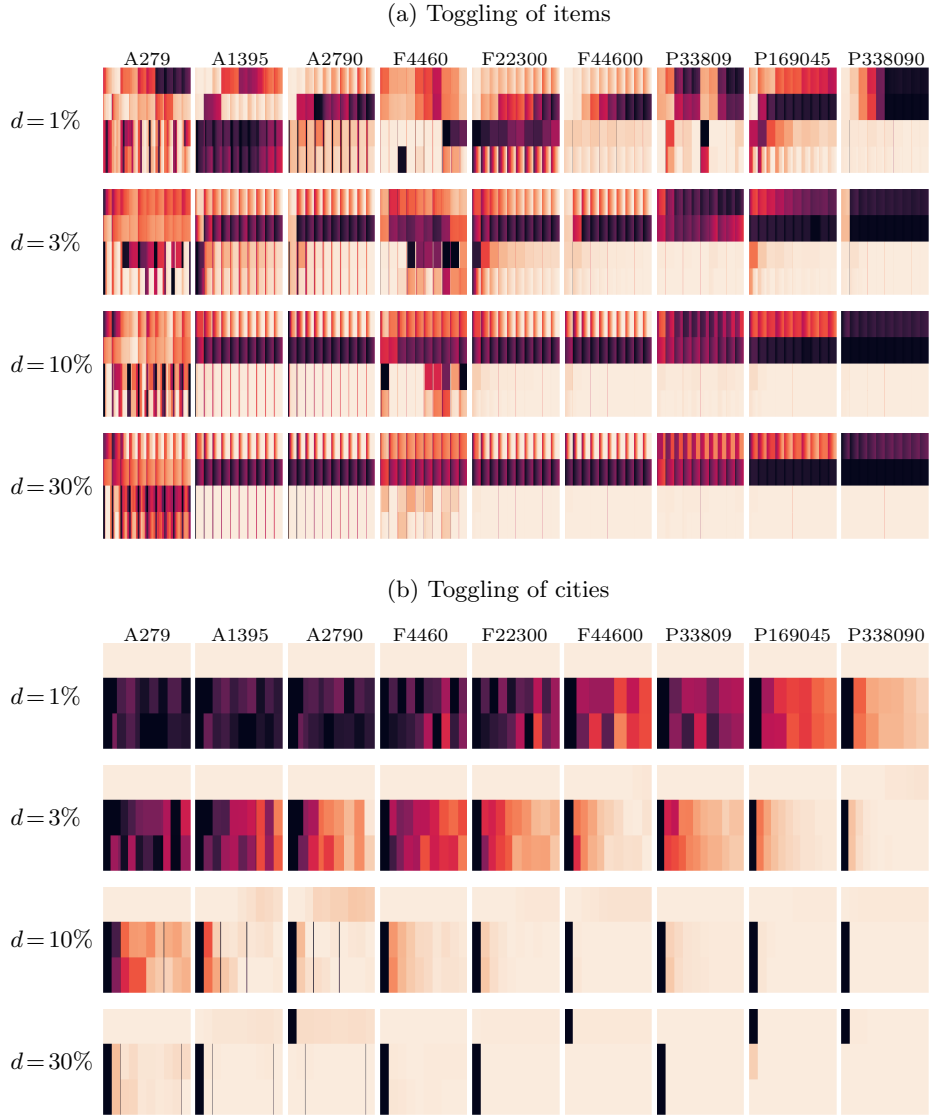

%\centering
%\setlength{\fboxsep}{0pt}\setlength{\fboxrule}{1pt}\fbox{\includegraphics[width=39mm,trim=58 38 46 42,clip]{heatmaps_multiMutations/fnl4461_n4460_bounded-strongly-corr_01_items_k=30.png}}
% \hm{fnl4461_n4460_bounded-strongly-corr_01_items_k=1.png}
% \hm{fnl4461_n4460_bounded-strongly-corr_01_items_k=3.png}
% \hm{fnl4461_n4460_bounded-strongly-corr_01_items_k=10.png}
% \hm{fnl4461_n4460_bounded-strongly-corr_01_items_k=30.png}\\
% \hmFour{fnl4461_n4460_bounded-strongly-corr_01_items_k}\\
\hspace{50mm}(a) Toggling of items\vspace{2mm}\\%\line(1,0){347}\\
\hspace*{15mm}{\scriptsize \asmall \hspace{3.5mm} \amedium \hspace{3mm} \alarge \hspace{2.mm} \fsmall \hspace{2.5mm} \fmedium \hspace{2mm} \flarge \hspace{1.mm} \psmall \hspace{1mm} \pmedium \hspace{0.3mm} \plarge}\\
\raisebox{6mm}{\footnotesize$d\hspace{-0.5mm}=\hspace{-0.5mm}1\%$}\hspace{3mm}\hmNine{items}{1}\\
\raisebox{6mm}{\footnotesize$d\hspace{-0.5mm}=\hspace{-0.5mm}3\%$}\hspace{3mm}\hmNine{items}{3}\\
\raisebox{6mm}{\footnotesize$d\hspace{-0.5mm}=\hspace{-0.5mm}10\%$}\hspace{1.4mm}\hmNine{items}{10}\\
\raisebox{6mm}{\footnotesize$d\hspace{-0.5mm}=\hspace{-0.5mm}30\%$}\hspace{1.4mm}\hmNine{items}{30}\vspace{2mm}\\
%%%%%%%%
\hspace*{50mm}(b) Toggling of cities\vspace{2mm}\\%\line(1,0){347}\\
\hspace*{15mm}{\scriptsize \asmall \hspace{3.5mm} \amedium \hspace{3mm} \alarge \hspace{2.mm} \fsmall \hspace{2.5mm} \fmedium \hspace{2mm} \flarge \hspace{1.mm} \psmall \hspace{1mm} \pmedium \hspace{0.3mm} \plarge}\\
\raisebox{6mm}{\footnotesize$d\hspace{-0.5mm}=\hspace{-0.5mm}1\%$}\hspace{3mm}\hmNine{cities}{1}\\
\raisebox{6mm}{\footnotesize$d\hspace{-0.5mm}=\hspace{-0.5mm}3\%$}\hspace{3mm}\hmNine{cities}{3}\\
\raisebox{6mm}{\footnotesize$d\hspace{-0.5mm}=\hspace{-0.5mm}10\%$}\hspace{1.4mm}\hmNine{cities}{10}\\
\raisebox{6mm}{\footnotesize$d\hspace{-0.5mm}=\hspace{-0.5mm}30\%$}\hspace{1.4mm}\hmNine{cities}{30}\\
%%%%%%%%
\vspace{-2mm}\caption{DynTTP results of seven algorithms on 72 scenarios, and for a duration of 10 epochs each. 
The performance has been normalised across all algorithms (within each individual epoch and w.r.t. to all objective scores seen in this epoch) to allow for a relative performance comparison. 
\textbf{Part (a)} Toggling of items: \bitflip (1st row), \rea (2nd row), \packit (3rd row), \packitbitflip (4th row). 
\textbf{Part (b)} Toggling of cities: \insertion (1st row), \clk (2nd row), \clkinsertion (3rd row). The lighter the colour, the higher the performance. 
\textbf{Example:} based on the heatmap in the top right corner of Part (a) (i.e., \plarge and $d=1\%$), we can see that all algorithms perform comparatively at first, but then \packit and \packitbitflip gradually perform better and better than the other two approaches.}
\label{fig:heatmapsAll}
\end{figure}

Figure~\ref{fig:heatmapsAll} shows all the results. While we will dig deeper in our following statistical analyses, it is obvious that some situations are ``very clear''. For example, for the \emph{Toggling Items} scenarios and for \plarge (rightmost column of heatmaps), the bottom two approaches always perform best (these are \packit and \packitbitflip, which both start from scratch) as they have the brightest colours. In contrast to this, the approaches that start from scratch for the \emph{Toggling Cities} scenarios (with $d=1\%$) perform the worst. We can also see that, as we move from left to right in this matrix of heatmaps, that patterns change and thus the ranking of the algorithms change -- this is important for anyone making decisions: based on this qualitative analysis, it is already clear that different situations require conceptually different approaches.

\paragraph{Statistical Analyses.} Given our high-dimensional dataset, we can cut through it in various ways and compare the performance of the different algorithms under various conditions. As inputs to these tests, we use the averaged (normalised) performance in two ways: (1) the final performance at the end of an epoch (END), and (2) the area-under-the-curve of the objective scores in an epoch (AUC), to consider the convergence speed as well as quality.

We employ the pairwise, one-sided Mann-Whitney~U~test as a ranksum test that does not rely on normal-distributed data. Again, we will limit ourselves to interesting cases, and to those with p-values of less than 5\%. 
%[remember: the performance of toggle-cities and toggle-items are not compatible in the normalisation]

In the following, we first provide a global picture, and then slice through the data along the four amounts of disruption $d$ and along the nine instances. This results in a total of 650 pair-wise statistical tests. Our summary: %In summary, we find the following:

% #optimisation = ["viaBitflip", "viaStructuralDiversity", "viaPackIterative", "viaPackIterativeAndBitflip"]
% #optimisation = ["viaJump", "viaCLK", "viaCLKAndJump"]
\vspace{-1mm}\begin{enumerate}

\item \emph{Global level (aggregated over all epochs, instances and values of $d$)}. 
(1) Toggling Items: Starting from scratch is preferred, as \packitbitflip outperforms \packit, which outperforms \bitflip and that in turn outperforms \rea. 
(2) Toggling Cities: Recovering is preferred here, as \insertion outperforms both \clk and \clkinsertion, which perform comparably. One of the reasons for this is that we do not optimise the packing in this case, and when we start from scratch the TSP is changed and the packing no longer is adequate with it. Moreover, the heuristics here are problem-independent, whereas \packit is TTP-specific.
% GLOBAL
% items
% AUC
% [[0.5 1.  0.  0. ]
%  [0.  0.5 0.  0. ]
%  [1.  1.  0.5 0. ]
%  [1.  1.  1.  0.5]]
% Final
% [[0.5   1.    0.006 0.   ]
%  [0.    0.5   0.    0.   ]
%  [0.994 1.    0.5   0.   ]
%  [1.    1.    1.    0.5  ]]
% cities
% AUC
% [[0.5   1.    1.   ]
%  [0.    0.5   0.597]
%  [0.    0.403 0.5  ]]
% Final
% [[0.5  1.   1.  ]
%  [0.   0.5  0.31]
%  [0.   0.69 0.5 ]]

\item \emph{Disruptions $d$ (aggregated over all epochs and instances)}:
(1) Toggling Items: We now see a few changes in the ranking when d=1\%, i.e., \bitflip now outperforms \packit as well (in addition to \rea, but only in case of AUC) possibly because the change to the instance is very minor and recovering is better than risking the start from a solution from scratch that is of unknown quality. However, when $d \geq 3\%$ the disruption seems too big for \bitflip to recover, and \packit again outperforms \bitflip.
% items
% AUC
% [[0.501 0.985 0.971*** 0.005]
%  [0.015 0.501 0.297 0.   ]
%  [0.029 0.704 0.501 0.   ]
%  [0.995 1.    1.    0.501]]
% Final
% [[0.501 0.999 0.792 0.   ]
%  [0.001 0.501 0.005 0.   ]
%  [0.209 0.995 0.501 0.   ]
%  [1.    1.    1.    0.501]]

% items
% AUC
% [[0.501 1.    0.    0.   ]
%  [0.    0.501 0.    0.   ]
%  [1.    1.    0.501 0.   ]
%  [1.    1.    1.    0.501]]
% Final
% [[0.501 1.    0.11  0.   ]
%  [0.    0.501 0.    0.   ]
%  [0.89  1.    0.501 0.   ]
%  [1.    1.    1.    0.501]]
 
%  items
% AUC
% [[0.501 1.    0.    0.   ]
%  [0.    0.501 0.    0.   ]
%  [1.    1.    0.501 0.   ]
%  [1.    1.    1.    0.501]]
% Final
% [[0.501 1.    0.    0.   ]
%  [0.    0.501 0.    0.   ]
%  [1.    1.    0.501 0.   ]
%  [1.    1.    1.    0.501]]
 
%  items
% AUC
% [[0.501 1.    0.    0.   ]
%  [0.    0.501 0.    0.   ]
%  [1.    1.    0.501 0.   ]
%  [1.    1.    1.    0.501]]
% Final
% [[0.501 1.    0.019 0.   ]
%  [0.    0.501 0.    0.   ]
%  [0.981 1.    0.501 0.   ]
%  [1.    1.    1.    0.501]]

(2) Toggling Cities: We observe only a single change from the global picture, i.e, \clkinsertion outperforms the other two when $d=30\%$ (END).% is considered. %$\rightarrow$  \mw{Possible reasons: ...}

% cities
% %crash due to same arrays
% cities
% AUC
% [[0.501 1.    1.   ]
%  [0.    0.501 0.495]
%  [0.    0.506 0.501]]
% Final
% [[0.501 1.    1.   ]
%  [0.    0.501 0.538]
%  [0.    0.463 0.501]]
% cities
% AUC
% [[0.501 1.    1.   ]
%  [0.    0.501 0.181]
%  [0.    0.819 0.501]]
% Final
% [[0.501 1.    1.   ]
%  [0.    0.501 0.495]
%  [0.    0.506 0.501]]
% cities
% AUC
% [[0.501 1.    1.   ]
%  [0.    0.501 0.786]
%  [0.    0.215 0.501]]
% Final
% [[0.501 0.084 0.   ]
%  [0.917 0.501 0.003]
%  [1.    0.997 0.501]]

\item \emph{Instances (aggregated over all epochs and values of $d$)}. 
(1) Toggling Items: \asmall is a surprise in that \rea outperforms all three in AUC; % (and two of them in END); 
for the next four larger instances (up to \fmedium) this does not hold anymore and many comparisons are insignificant; for the largest five instances, the global picture holds and \packitbitflip is the always significantly better. 
(2) Toggling Cities: There is never any deviation from the global picture, except for \plarge, where there are not significant differences of all three approaches when the final performance (END) is considered. 

\end{enumerate}

\noindent To sum up, we observe that disruptive events have a major impact on the performance of the algorithms, which in turn significantly affect the relative rankings.

\section{Outlook}

%While many real-world problems are inherently dynamic and contain multiple interacting components, most research on dynamic problems appears to focus on single-component problems. To address this, we have introduced the DynTTP as a dynamic variant of the multi-component Travelling Thief Problem. We have evaluated the performance of seven TTP approaches in 72 different scenarios, in which we varied the instances, the amount of disruption and the type of disruption. We have found that, depending on the scenario, it is better to either recover from the previous solution, or to start from scratch.

Future research can go into many different directions. For example, none of the approaches was properly problem-specific, and only \rea employed a population. 
Even though \rea did not perform well, we intend to explore how to adapt its diversity mechanism to entire TTP solutions. In addition, we envision multi-objective approaches and co-evolutionary approaches that will balance solution quality while preparing for disruptive events of unknown type and magnitude.

\vspace{2mm}
\noindent\textit{Acknowledgement.} This work has been supported by the Australian Research Council through grants DP160102401 and DP200102364.
% \section*{Acknowledgements}

% (removed for double-blind review)

\bibliographystyle{splncs04}
\bibliography{mybibliography,donteditme_springer_thief}

\begin{thebibliography}{10}
\providecommand{\url}[1]{\texttt{#1}}
\providecommand{\urlprefix}{URL }
\providecommand{\doi}[1]{https://doi.org/#1}

\bibitem{DBLP:conf/cec/Ameca-AlducinHB18}
Ameca{-}Alducin, M.Y., Hasani{-}Shoreh, M., Blaikie, W., Neumann, F.,
  Mezura{-}Montes, E.: A comparison of constraint handling techniques for
  dynamic constrained optimization problems. In: {IEEE} Congress on
  Evolutionary Computation. pp.~1--8. {IEEE} (2018)

\bibitem{DBLP:conf/evoW/Ameca-AlducinHN18}
Ameca{-}Alducin, M.Y., Hasani{-}Shoreh, M., Neumann, F.: On the use of repair
  methods in differential evolution for dynamic constrained optimization. In:
  21st International Conference EvoApplications. vol. 10784, pp. 832--847.
  Springer (2018)

\bibitem{chainedLK03applegate}
Applegate, D., Cook, W.J., Rohe, A.: Chained {L}in-{K}ernighan for large
  traveling salesman problems. INFORMS Journal on Computing  \textbf{15}(1),
  82--92 (2003)

\bibitem{DBLP:journals/corr/abs-2002-06766}
Assimi, H., Harper, O., Xie, Y., Neumann, A., Neumann, F.: Evolutionary
  bi-objective optimization for the dynamic chance-constrained knapsack problem
  based on tail bound objectives  (2020),
  \url{https://arxiv.org/abs/2002.06766}

\bibitem{blank2017solvingBittp}
Blank, J., Deb, K., Mostaghim, S.: Solving the Bi-objective Traveling Thief
  Problem with Multi-objective Evolutionary Algorithms, pp. 46--60. Springer
  (2017)

\bibitem{DBLP:conf/cec/BonyadiMB13}
Bonyadi, M.R., Michalewicz, Z., Barone, L.: {The Travelling Thief Problem: The
  First Step in the Transition from Theoretical Problems to Realistic
  Problems}. In: {IEEE Congress on Evolutionary Computation}. pp. 1037~--~1044.
  IEEE (2013)

\bibitem{bonyadi2013travelling}
Bonyadi, M.R., Michalewicz, Z., Barone, L.: The travelling thief problem: The
  first step in the transition from theoretical problems to realistic problems.
  In: IEEE Congress on Evolutionary Computation. pp. 1037--1044. IEEE (2013)

\bibitem{bonyadi2014socially}
Bonyadi, M.R., Michalewicz, Z., Przybylek, M.R., Wierzbicki, A.: Socially
  inspired algorithms for the {TTP}. In: Genetic and Evolutionary Computation
  Conference. pp. 421--428. ACM (2014)

\bibitem{Bonyadi2019}
Bonyadi, M.R., Michalewicz, Z., Wagner, M., Neumann, F.: Evolutionary
  Computation for Multicomponent Problems: Opportunities and Future Directions,
  pp. 13--30. Springer (2019)

\bibitem{chagas2020nondominated}
Chagas, J.B.C., Blank, J., Wagner, M., Souza, M.J.F., Deb, K.: A non-dominated
  sorting based customized random-key genetic algorithm for the bi-objective
  traveling thief problem. CoRR  \textbf{abs/2002.04303} (2020),
  \url{https://arxiv.org/abs/2002.04303}

\bibitem{chand2016fast}
Chand, S., Wagner, M.: Fast heuristics for the multiple traveling thieves
  problem. In: Genetic and Evolutionary Computation Conference. pp. 293--300.
  GECCO '16, ACM (2016)

\bibitem{ReoptViaStructuralDiversity}
Doerr, B., Doerr, C., Neumann, F.: Fast re-optimization via structural
  diversity. CoRR  \textbf{abs/1902.00304} (2019),
  \url{http://arxiv.org/abs/1902.00304}

\bibitem{elyafrani2018hyperttp}
El~Yafrani, M., Martins, M., Wagner, M., Ahiod, B., Delgado, M., L{\"u}ders,
  R.: A hyperheuristic approach based on low-level heuristics for the
  travelling thief problem. Genetic Programming and Evolvable Machines
  \textbf{19}(1),  121--150 (2018)

\bibitem{faulkner2015approximate}
Faulkner, H., Polyakovskiy, S., Schultz, T., Wagner, M.: Approximate approaches
  to the traveling thief problem. In: Genetic and Evolutionary Computation
  Conference. pp. 385--392. ACM (2015)

\bibitem{ApproximateApproaches}
Faulkner, H., Polyakovskiy, S., Schultz, T., Wagner, M.: Approximate approaches
  to the traveling thief problem. In: Genetic and Evolutionary Computation
  Conference. pp. 385--392. ACM (2015)

\bibitem{herring2020dynamic}
Herring, D., Kirley, M., Yao, X.: Dynamic multi-objective optimization of the
  travelling thief problem. CoRR  \textbf{abs/2002.02636} (2020)

\bibitem{klamroth2016interwoven}
Klamroth, K., Mostaghim, S., Naujoks, B., Poles, S., Purshouse, R., Rudolph,
  G., Ruzika, S., Sayın, S., Wiecek, M.M., Yao, X.: Multiobjective
  optimization for interwoven systems. J. of Multi-Criteria Decision Analysis
  \textbf{24}(1-2),  71--81 (2017)

\bibitem{Martins2017ttpeda}
Martins, M.S.R., El~Yafrani, M., Delgado, M.R.B.S., Wagner, M., Ahiod, B.,
  L\"{u}ders, R.: Hseda: A heuristic selection approach based on estimation of
  distribution algorithm for the travelling thief problem. In: Genetic and
  Evolutionary Computation Conference. p. 361–368. ACM (2017)

\bibitem{mei2014interdependence}
Mei, Y., Li, X., Yao, X.: On investigation of interdependence between
  sub-problems of the {TTP}. Soft Computing  \textbf{20}(1),  157--172 (2014)

\bibitem{namazi2019cooperative}
Namazi, M., Sanderson, C., Newton, M.A.H., Sattar, A.: A cooperative
  coordination solver for travelling thief problems. ArXiv e-prints  (2019)

\bibitem{neumann2017ttpPTAS}
Neumann, F., Polyakovskiy, S., Skutella, M., Stougie, L., Wu, J.: A fully
  polynomial time approximation scheme for packing while traveling. In:
  Algorithmic Aspects of Cloud Computing. pp. 59--72. Springer (2019)

\bibitem{BenchmarkPaper}
Polyakovskiy, S., Bonyadi, M.R., Wagner, M., Michalewicz, Z., Neumann, F.: A
  comprehensive benchmark set and heuristics for the traveling thief problem.
  In: Genetic and Evolutionary Computation Conference. pp. 477--484. ACM (2014)

\bibitem{DBLP:conf/ppsn/Roostapour0N18}
Roostapour, V., Neumann, A., Neumann, F.: On the performance of baseline
  evolutionary algorithms on the dynamic knapsack problem. In: Parallel Problem
  Solving from Nature. pp. 158--169 (2018)

\bibitem{DBLP:journals/corr/abs-1811-07806}
Roostapour, V., Neumann, A., Neumann, F., Friedrich, T.: Pareto optimization
  for subset selection with dynamic cost constraints. In: Proc. of {AAAI}. pp.
  2354--2361 (2019)

\bibitem{DBLP:journals/corr/abs-1806-08547}
Roostapour, V., Pourhassan, M., Neumann, F.: Analysis of evolutionary
  algorithms in dynamic and stochastic environments. In: Theory of Evolutionary
  Computation, pp. 323--357. Springer (2020)

\bibitem{DBLP:journals/algorithmica/ShiSFKN19}
Shi, F., Schirneck, M., Friedrich, T., K{\"{o}}tzing, T., Neumann, F.:
  Reoptimization time analysis of evolutionary algorithms on linear functions
  under dynamic uniform constraints. Algorithmica  \textbf{81}(2),  828--857
  (2019)

\bibitem{wagner2016stealing}
Wagner, M.: Stealing items more efficiently with ants: A swarm intelligence
  approach to the travelling thief problem. In: Swarm Intelligence. pp.
  273--281. Springer (2016)

\bibitem{wagner2018casestudy}
Wagner, M., Lindauer, M., M{\i}s{\i}r, M., Nallaperuma, S., Hutter, F.: A case
  study of algorithm selection for the traveling thief problem. Journal of
  Heuristics  \textbf{24}(3),  295--320 (2018)

\bibitem{wu2018evolutionary}
Wu, J., Polyakovskiy, S., Wagner, M., Neumann, F.: Evolutionary computation
  plus dynamic programming for the bi-objective travelling thief problem. In:
  Genetic and Evolutionary Computation Conference. pp. 777--784. ACM (2018)

\bibitem{wu2017ttpexact}
Wu, J., Wagner, M., Polyakovskiy, S., Neumann, F.: Exact approaches for the
  travelling thief problem. In: Simulated Evolution and Learning. pp. 110--121.
  Springer (2017)

\bibitem{Wuijts2019ttpinvest}
Wuijts, R.H., Thierens, D.: Investigation of the traveling thief problem. In:
  Genetic and Evolutionary Computation Conference. p. 329–337. ACM (2019)

\bibitem{yafrani2017ttpemo}
Yafrani, M.E., Chand, S., Neumann, A., Ahiod, B., Wagner, M.:
  Multi-objectiveness in the single-objective traveling thief problem. In:
  Genetic and Evolutionary Computation Conference Companion. pp. 107--108. ACM
  (2017)

\bibitem{elyafrani2018ttplandscape}
Yafrani, M.E., Martins, M.S.R., Krari, M.E., Wagner, M., Delgado, M.R.B.S.,
  Ahiod, B., L\"{u}ders, R.: A fitness landscape analysis of the travelling
  thief problem. In: Genetic and Evolutionary Computation Conference. p.
  277–284. ACM (2018)

\bibitem{Zouari2019antstpp}
Zouari, W., Alaya, I., Tagina, M.: A new hybrid ant colony algorithms for the
  traveling thief problem. In: Genetic and Evolutionary Computation Conference
  Companion. p. 95–96. ACM (2019)

\end{thebibliography}

\end{document}